\documentclass[sigconf, numbers]{acmart}

\usepackage[utf8]{inputenc} 
\usepackage[T1]{fontenc}    
\usepackage{hyperref}       
\usepackage{url}            
\usepackage{booktabs}       
\usepackage{nicefrac}       
\usepackage{microtype}      
\usepackage{bookmark}
\usepackage{graphicx}
\usepackage{amsmath,amsfonts, bm}
\usepackage{footnote}
\usepackage{caption}
\usepackage{subcaption}
\usepackage{siunitx}
\usepackage{mathtools}
\usepackage{hyperref}
\usepackage{appendix}

\usepackage{stfloats}

\DeclareMathOperator{\Exc}{\mathbb{E}}
\newcommand{\norm}[1]{\left\lVert#1\right\rVert}

\DeclarePairedDelimiter\abs{\lvert}{\rvert}%

\makeatletter
\let\oldabs\abs
\def\abs{\@ifstar{\oldabs}{\oldabs*}}
\let\oldnorm\norm
\def\norm{\@ifstar{\oldnorm}{\oldnorm*}}
\makeatother

\setcopyright{acmcopyright}
\copyrightyear{2020}
\acmYear{2020}
\acmDOI{}

\acmConference[DTMBio 2020]{ACM 14th International Workshop on Data and Text Mining in Biomedical Informatics}
{19-23 October 2020}{Galway, Ireland}

\begin{document}
\pagenumbering{arabic}                
\pagestyle{plain}

\title{Mapping of Lesion Images to Somatic Mutations}

\author{Rahul Mehta}
\email{mehta5@uic.edu}
\affiliation{\institution{University of Illinois at Chicago}
\city{Chicago}
 \state{Illinois}
 \country{USA}
 }
 \author{Yang Lu}
\email{ylu@mdanderson.org}
\affiliation{\institution{University of Texas MD Anderson Cancer Center}
\city{Houston}
 \state{Texas}
 \country{USA}
 }
\author{Muge Karaman}
\email{mkaraman@uic.edu}
\affiliation{\institution{University of Illinois at Chicago}
\city{Chicago}
 \state{Illinois}
 \country{USA}
 }


\begin{abstract}
    Medical imaging is a critical initial tool used by clinicians to determine a patient's cancer diagnosis, allowing for faster intervention and more reliable patient prognosis.  At subsequent stages of patient diagnosis, genetic information is extracted to help select specific patient treatment options.  As the efficacy of cancer treatment often relies on early diagnosis and treatment, we build a deep latent variable model to determine patients' somatic mutation profiles based on their corresponding medical images.  We first introduce a point cloud representation of lesions images to allow for invariance to the imaging modality.  We then propose, LLOST, a model with dual variational autoencoders coupled together by a separate shared latent space that unifies features from the lesion point clouds and counts of distinct somatic mutations.  Therefore our model consists of three latent space, each of which is learned with a conditional normalizing flow prior to account for the diverse distributions of each domain.  We conduct qualitative and quantitative experiments on de-identified medical images from The Cancer Imaging Archive and the corresponding somatic mutations from the Pan Cancer dataset of The Cancer Genomic Archive.  We show the model's predictive performance on the counts of specific mutations as well as it's ability to accurately predict the occurrence of mutations.  In particular, shared patterns between the imaging and somatic mutation domain that reflect cancer type.  We conclude with a remark on how to improve the model and possible future avenues of research to include other genetic domains.
\end{abstract}


\maketitle

\section{Introduction}\label{intro}

Targeted cancer therapies, those based on identifying the genetic makeup of a legion, are often more effective and can incur fewer side effects than traditional therapies \cite{murdoch2008will}. Unfortunately, the efficacy of these treatments is highly dependent on early detection and treatment and so delayed turnaround time of genetic analyses can result in negative consequences for survival rate. To bypass the time-consuming process of biopsy and genetic analysis, computational models can exploit quantitative and qualitative information derived from medical images of lesions to obtain cancer imaging features\footnote{For the remainder of the paper we refer them as imaging features} that can predict genetic markers present in the individual patient.  These genetic markers can then be leveraged to project patient prognosis and determine targeted treatment options. 

A design challenge in computational models that predict genetic markers from imaging features is the inherent genetic heterogeneity of cancerous lesions.  Current models overcome this limitation by enforcing imaging features to discriminate between a specific set or cluster of genetic markers \cite{lambin2012radiomics, gevaert2012non, nair2012prognostic}.  This design choice, however, incurs false positives in downstream tasks for patient prognosis \cite{bai2016imaging}.  While a model may correctly predict the presence of a genetic marker, it is often the confluence of many genes that influence cancer progression \cite{pharoah2004association}.  As cancerous lesions occur as a result of a combination of genetic and epigenetic changes within a patient, imaging features must predict all possible genetic markers \cite{martincorena2017universal} to accurately assess the correct treatment options for a patient. 

The idea of exploiting computational cancer imaging features from lesion images for patient analysis goes back decades \cite{wang2012machine}.  The general framework for the task of prediction using imaging features is posited as: given a black box model (e.g., a neural network) that converts a medical image into a set of imaging features, we can map the imaging features onto a function (e.g., Softmax) which learns parameters that optimizes the prediction accuracy of a set of labels (genetic markers) of interest.  Following the same paradigm, we would like our model to map the image of a lesion onto a patient's full somatic mutation profile.  The learning task is unfortunately, impeded by the overwhelming size of the output space, i.e., the somatic mutation profile.

Our goal of mapping a lesion image to a somatic mutation profile is analogous to many methods in domain mapping \cite{sun2016return} where a high-dimensional input is mapped to a high-dimensional output as in text translation, image to image translation, and image captioning.  Current approaches in these applications have leveraged deep latent variable models \cite{goodfellow2014generative,rezende2015variational,dinh2014nice} to enforce a shared latent space \cite{wang2016learning} or a cyclical structure \cite{zhu2017unpaired}.  The main objective in these models is to minimize the loss when translating between domains such that model can recover the data from the original domain, i.e. domain A to domain B back to domain A.  The key advantage of deep latent variable models is that the neural architecture can conserve complex correlations among different domains by constraining the loss functions.  For example, in image captioning, the neural network architecture can conserve the shape features of a mountain across populations (height, peak, width), while also changing these correlations respective to mutable semantic characteristics such as "snowy" or "volcano."  We can apply the same concepts in the biological domain, however, we must consider two specific challenges:

1) Many models featuring cancer imaging features use a single slice of a lesion, however, we cannot follow the same assumption since the exact location of the lesion biopsies are unknown.  Additionally, lesion images come from multiple imaging modalities such as computerized tomography (CT) or magnetic resonance imaging (MRI).  Therefore, our model must incorporate all of the lesion slices during inference, while also remaining invariant to the imaging domain.

2) Somatic mutation datasets, like most genetic datasets, are discrete and high-dimensional.  Complexity is further increased due to sparsity, that is, the data are characterized by few frequently occurring mutations and a long-tail of rare mutations.  Therefore, our model must use a function that mitigates the underfitting of the data.

In this paper, we construct, Lesion Point Cloud to Somatic Mutations, LLOST, with dual variational autoencoders (VAE) \cite{kingma2013auto} where each encoder/decoder architecture represents the domains of interest: the lesion image and the somatic mutations.  For the lesion VAE, we use a point-cloud encoder/decoder architecture, which allows us to be invariant to the imaging domain.  For the somatic mutation VAE, we use a Negative-Binomial likelihood to model the sparsity and high dimensionality of the dataset.  The two VAEs are coupled together with shared latent space modeled by a a single invertible neural network conditioned on the cancer type.  Each VAE also consists of its own domain specific latent space with conditional normalizing flows priors as a way to model the complexity of the two very different distributions \cite{dinh2016density,kingma2018glow}.  The main idea is to use one domain to generate the other by using a series of normalizing flows conditioned on features learned in the shared latent space.  Hence, by virtue of the LLOST's framework, we can transfer features from the lesion domain via the shared latent space to create a mutation specific latent space from the conditional prior.  The concatenation of the two latent spaces then generates a prediction of how many times each distinct gene is mutated i.e., a full somatic mutation profile of a patient. 

Our choice of utilizing somatic mutation profiles in comparison to other cancer genetic markers is that the somatic mutation profiles provide two immediate applications for clinicians.  One use for clinicians is the identification of co-occurring somatic mutations of interest for early targeted treatment such as immunotherapies \cite{tran2017final}.  The other is to predict a patient's tumor mutational load (TML), a sum of the total number of mutations in a lesion, which current research has proposed as a potential biomarker for determining patient prognosis and sensitivity to targeted treatments \cite{samstein2019tumor}.  

To analyze LLOST, we use the somatic mutation and lesion imaging data from The Cancer Genomic Archive and The Cancer Imaging Archive, respectively, for inference and prediction.   We analyze the predicted somatic mutation profiles using perplexity and distance measurements.  A ramification of our work is a better assessment of treatment options for patients at earlier stages of diagnosis, as well as the possibility of further downstream tasks such as prognosis of patients using tumor mutational load.  

\section{Background}

In this section we discuss the necessary background on the components of our model: point cloud data, high dimensional discrete data, and normalizing flows.  We use uppercase notation to describe the lesion and mutation domains, $M$ and $I$, respectively.  We use bold lowercase notation to describe latent space parameters, where the subscripts depict a specific domain ($M$,$I)$ or sample ($n$), and unbolded lowercase notation to denote values assumed by the variables.

\subsection{Point Clouds} \label{feats}

Point clouds have gained traction in the analysis of 3D objects with the increasing availability of 3D sensors and acquisition technologies.  Point clouds are particularly amenable in describing arbitrary shapes as each point is simply a $(x,y,z)$ coordinate in Euclidean space, thereby offering a compact representation of surface geometry.  Recent deep learning models are built upon the nascent ideas introduced in PointNet \cite{qi2017pointnet}, which operate directly on the point cloud, and therefore follow a more data-driven approach of extracting features.  Several studies have extended PointNet for different applications in shape completions, 3D segmentation, and 3D classification, and point cloud generation \cite{brock2016generative}.  The synergistic component within these models is the feature vector that aggregates global and local features of the point cloud.  Depending on the neural network architecture, the feature vectors can include descriptions of shape, volume, surface topology, object geometry, and the relationship between individual points.  

\subsection{Normalizing Flows}

Normalizing flows (NF) \cite{dinh2016density, rezende2015variational} is a type of likelihood based deep latent variable model that aims to map a simple base density, $p(\epsilon)$, to a complex density, $p(z)$, through several invertible parametric transformations with tractable Jacobians. An example of such a density is:

 \begin{equation}
  \begin{gathered}
      p_{\bm{z}}(\bm{z};\theta) = p_{\mathbf{\epsilon}}(f_{\theta}(\bm{z}))\abs{\text{det}\frac{df_{\theta}(\bm{z})}{dz}}
  \end{gathered}
  \label{eq:nf}
\end{equation} 

where $f: \mathcal{R}^{D} \mapsto \mathcal{R}^{D}$ is an invertible function with parameters $\theta$.   

The main advantage of NF is its inherent invertible construction that allows training in the forward and the reverse directions.  Recently, Ardizzone et al. \cite{ardizzone2018analyzing} exploited this mechanism for inverse sampling problems when both distributions are arbitrarily complex.  The only limitation is that the model has to sample from both distributions and evaluate the forward and and backward processes.  These requirements are easily satisfied using neural network architecture based on current literature such as in RealNVP \cite{dinh2016density} and Autoregressive Flows \cite{kingma2016improved}.  Since our data comes from two complex distributions, NF provide an intuitive method to move between distributions and uncover distinct properties of the latent space.

\subsection{High Dimensional Discrete Data}

Parameter estimation of discrete count data is often done using latent variable models (LVM), $p(x) =\int_{z}p(x|z)p(z)dz$, where $p(x)$ is the distribution of the original data, $p(z)$ is the prior for the latent distribution and $p(x|z)$ is some probability distribution.  A caveat of LVMs is that the practitioner must determine which distributions to use for $p(z)$ and $p(x|z)$ to correctly capture the underlying distribution of the data \cite{krishnan2017challenges}. 

The Negative-binomial (NB) distribution has emerged as the probability distribution of choice to estimate count data often seen in genetic datasets \cite{bailey2018comprehensive}.  We exploit the inference methods of a VAE to learn the parameters of the NB distribution,  \cite{DBLP:journals/corr/abs-1905-00616} by rewriting the VAE generative process as:

\begin{equation}
  \begin{gathered}
      \mathbf{z_M} \sim \mathcal{N}(0,1); \mathbf{r_M} \sim exp(f_{\theta^r}(\mathbf{z_M})); \\ \mathbf{p_M} \sim \frac{1}{1+exp(-f_{\theta^p}(\mathbf{z_M}))}; M_n \sim NB(\mathbf{r_M}, \mathbf{p_M})
  \end{gathered}
  \label{eq:nb_reg}
\end{equation} 

We can also use the the NB parameters to generate binary labels, i.e., we would instead predict if a specific somatic mutation occurs in a patient rather than the number of times it occurs.  The likelihood follows a Bernoulli distribution:

\begin{equation}
  \begin{gathered}
      M_{n}^{b} \sim Bernoulli(1- (1-\mathbf{p_M})^{\mathbf{r_M}})
  \end{gathered}
  \label{eq:nb_bin}
\end{equation} 

An important design choice in LVMs is the choice of the distribution of the prior. The prior for the VAE in Equation \ref{eq:nb_reg} uses univariate Gaussian prior for the latent space.  This can become incongruous as datasets become more complex \cite{krishnan2017challenges}. There are several priors available to us to increase the effectiveness of LVMs.  What we need to acknowledge, is the ability to perform efficient inference and the flexibility to learn the distributions of two very different datasets.

\begin{figure}%
\centering
\includegraphics[width=1\linewidth]{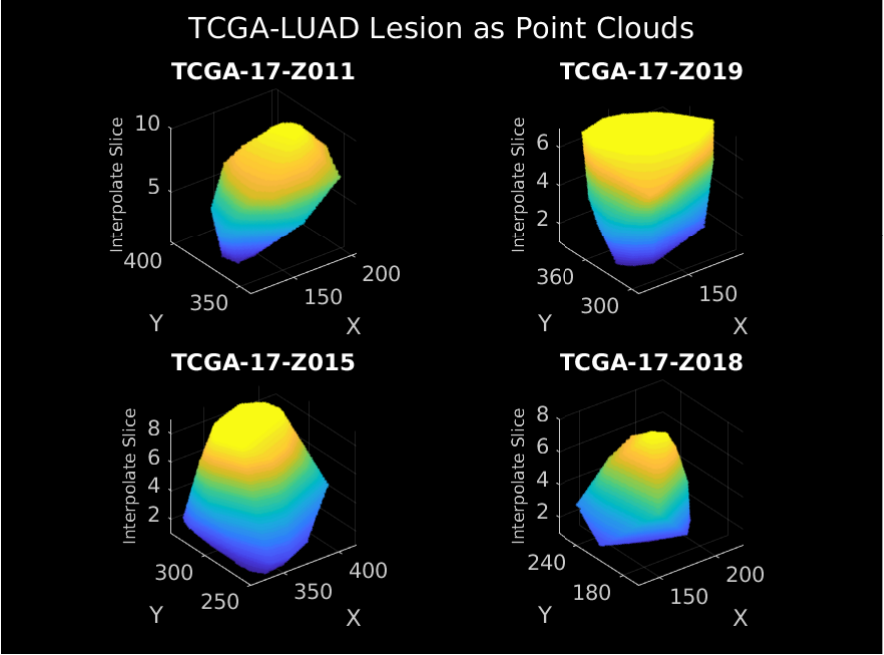}%
\caption{}
\label{fig:interpolated}%
\caption{Shows examples of lung cancer lesions from different patients as points clouds respective of the axial plane.  The points were generated by interpolating along with z-axis, with respect to the individual lesion slices.}
\end{figure} 

\subsection{Related Work}\label{rel}

Early techniques for multimodal learning or domain mapping framed the learning paradigm as information retrieval, where the best domain match is optimized from a pool of embeddings in a shared latent space \cite{rasiwasia2010new}. These are now superceded by deep latent variable models (DLVM) such as VAEs or General Adversial Networks (GAN) \cite{goodfellow2014generative} that scale to larger dataset, while still employing a strategy of sharing a latent space.  A popular choice for multimodal learning in DLVMs is the conditional variational autoencoder (CVAE), where a shared latent space is generated by concatenating the observed labels with the latent space.  

A limitation of a CVAE is that the learned latent space tends to encourage a distribution that encompasses dominant patterns of the data.  Since labels are not available during test time, a CVAE model will only predict labels that approximate the original training distribution.  Many authors circumvent this issue in a CVAE via modification of the prior for the latent space \cite{tomczak2017vae, kingma2016improved}.  While the results of various CVAE models are impressive, sharing the latent space removes domain specific features \cite{sohn2015learning} and leads to limited diversity in the latent structure.  To counteract this issue, Mahajan et al. \cite{mahajan2020latent} created a separate shared latent space along with a domain specific latent space to generate a more diverse latent structure for image captioning.  Similarly, Xie et al. \cite{xie2019dual} used a separate embedding layer to better model semantically similar texts.  Inspired by these models, LLOST creates a distinct shared latent space augmented by conditioning it on the cancer type.

\section{Dataset}

We use the Pan Cancer Dataset from TCGA \cite{liu2018integrated}, which contains the unique somatic mutation profiles for 10295 patients.  We then find the corresponding lesion image(s) of each patient from TCIA database (from any modality).  If a cancer type had less than 10 patient samples we did not include it in our dataset.  The final dataset consists of 1342 patients from 18 cancer types each with a lesion image and a corresponding somatic mutation profile as shown in Table \ref{tab:cancer_result}. 

\subsection{Lesion as a Point Cloud}
 
A key decision in building our dataset concerned the context of how our model could extract diverse and specific information from the lesion image regardless of cancer type and image modality.  Currently, there is no explicit way to compare lesion images as they are highly irregular and heterogeneous among samples.  We propose to use point clouds as a way to model the intrinsic antisotropic and heterogeneous nature of lesions.  Point clouds offer a rich interpretation of lesions as discussed in Section \ref{feats}, and have several other attractive properties that pertain to the medical imaging dataset.  In particular, since point clouds are a set of un-ordered points, they are modality independent (CT or MRI) compared to a set of pixels where each pixel intensity depends on the imaging modality.  Furthermore, a point cloud lesion decreases the computational footprint since each lesion is now represented as a 2D matrix in comparison to a volumetric lesion which is a 3D matrix.

To create the lesion point clouds, we first extract the lesions from individual slices using the segmentation labels provided by TCIA.  If segmentation labels were not available, a radiologist delineated the lesions within the slice and then transform the lesions into their real-world coordinates using the information stored within their respective DICOM files.  Figure \ref{fig:interpolated} shows the voxels within a lesion volume are then interpolated to create a point cloud following the model in \cite{amidror2002scattered}. Finally, for a specific lesion point cloud $I_n$, each point in $I_n$ is a set of coordinates from $\mathbb{R}^3$ uniformly sampled from the surface of the full lesion point cloud.  

\subsection{Somatic Mutations Representation}

Somatic mutations are represented as a count matrix, $M$, where each row in $M$ is a patient, and each column is a distinct gene.  A single element in the matrix indicates the number of times a gene is mutated.  This is analogous to Bag of Words for text datasets \cite{blei2012probabilistic}.  In our case, the vocabulary, $V$, is the set of all the genes (21332), $N$ is the total number of samples, and $M_n$ is a vector of counts of the $n^{\text{th}}$ sample.  A binarized version of matrix, where each vector, $M_n$, indicates the occurrence of a matrix is used for the Bernoulli likelihood of our model.

\section{Model}

Given the above machinery, we represent our full dataset as $N$ patients, where each patient $n$ is associated with a point cloud $\mathbf{I_n}$, a count vector $\mathbf{M_n}$, and one hot vector label, $y_{n}$ that identifies the cancer type. Our goal is to learn the conditional distribution $p(M | I, y, \bm{z})$ where \bm{$z$} is a stochastic latent variable.  

\subsection{Lesion Point Cloud to Somatic Mutations}

\begin{figure*}
  \centering
  \includegraphics[width=.75\textwidth]{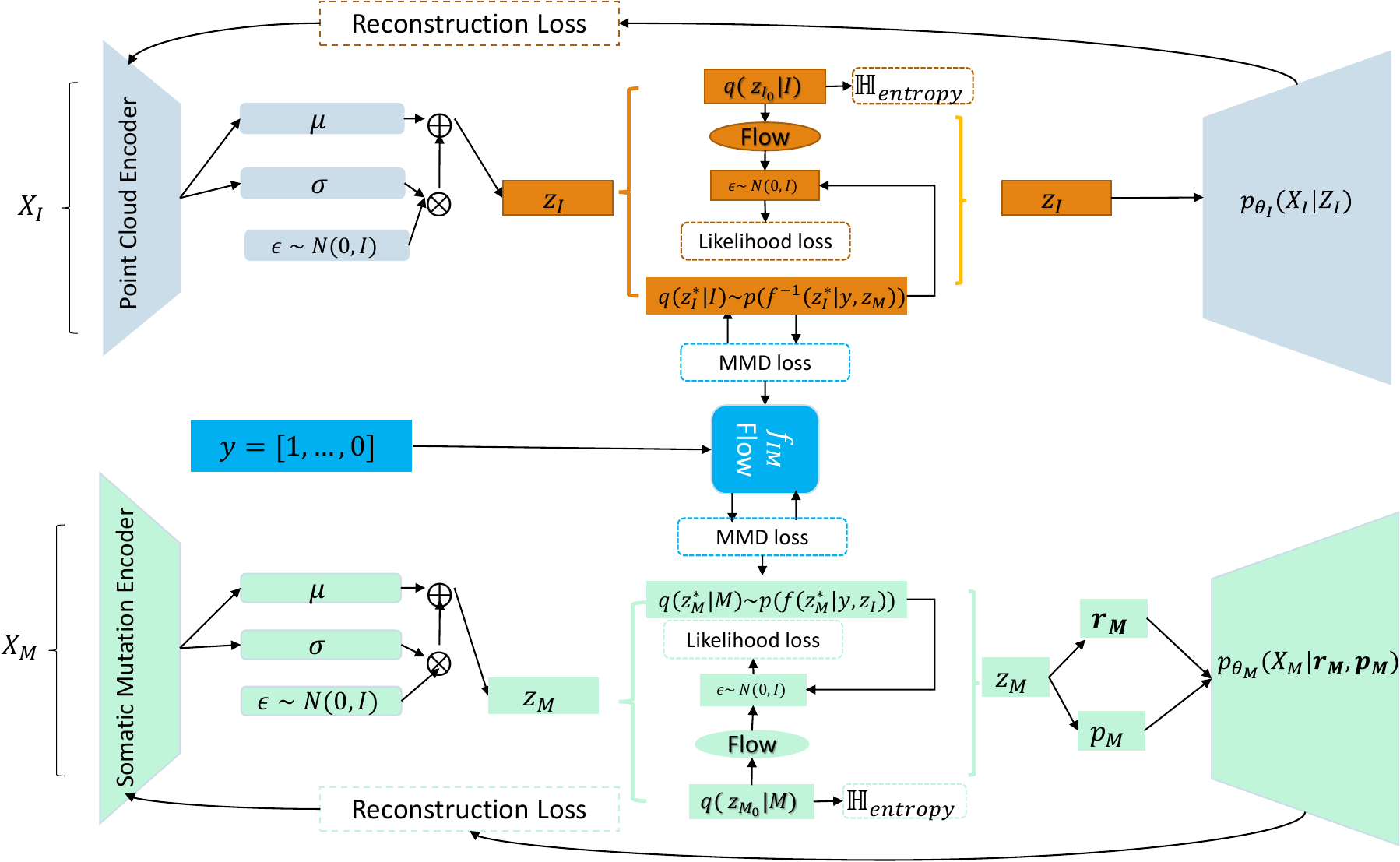}
  \caption{Model Architecture of LLOST.  During training, the approximate posterior distribution of the domain specific embedding tries to match the true posterior with a learnable prior conditioned on the shared latent spaces.  The shared latent space is trained by matching the distribution of the domain, so that it maps shared embeddings to domain specific embeddings.  The model is trained bidirectionally to maximize the ELBO, which is a sum of reconstruction loss, the KL divergence of the conditional NF, and the MMD loss of the shared latent space.  For clarity we drop the subscripts referring to the individual neural network parameters.}
    \label{fig:fullModel}
  
\end{figure*}

The Lesion Point Cloud to Somatic Mutation (LLOST) learns the conditional distribution $p(M | I, y, \bm{z})$ using two VAEs for each distinct domain identified as, $X_I$ (lesion) and $X_M$ (mutation).  Each VAE embeds the data into a lower dimensional latent space $\bm{z_I}$ and $\bm{z_M}$, respectively. Instead of using a single latent for each domain, we propose to use two domain specific latent spaces and a shared latent space.  We denote $\bm{z_{M_0}}$ for the mutation specific latent space, $\bm{z_{I_0}}$ for the lesion specific latent space, $\bm{z^{*}_{M}}$ for the shared mutation latent space, and $\bm{z^{*}_{I}}$ for the shared lesion latent space.  So $z_I = [\bm{z^{*}_{I}}, \bm{z_{I_0}}]$ and $z_M = [\bm{z^{*}_{M}}, \bm{z_{M_0}}]$.  The conditional distribution of the domains is then $p(M, I| y, \bm{z_{M_0}}, \bm{z^{*}_{M}}, \bm{z_{I_0}}, \bm{z^{*}_{I}})$ and is learned by approximating the latent variables using a variational posterior $q(\bm{z^{*}_{M}},\bm{z^{*}_{I}},\bm{z_{I_0}},\bm{z_{M_0}}|M,I,y)$.

Whereas the original VAE uses a fixed standard Gaussian prior for the latent space, we use a trainable prior via a conditional NF to model the complexity of the distributions of each domain \cite{ziegler2019latent}.  Having a trainable prior allows us to adapt the shape of the prior based on the data, and avoids fewer modeling assumptions \cite{rezende2015variational, sonderby2016ladder}.  Using Equation \ref{eq:nf}, our conditional NF prior for the image domain is constructed as $p_{\eta_{I}}(\bm{z_{I_0}}|\bm{z^{*}_{I}}) = p_{\mathbf{\epsilon}}(f_{\eta_{I}}(\bm{z_{I_0}}|\bm{z^{*}_{I}}))\abs{\text{det}\frac{df_{\eta_I}}{dz}}$, where $\eta_{I}$ are the parameters of the neural network.  This formulation allows the model to push the prior towards matching the approximate posterior during training.   

We take advantage of the conditional NF architecture again in the shared latent space, where the base distribution for the NF is now the latent space $q(\bm{z^{*}_{M}}|M)$.  This is modeled as a single invertible neural network, $f_{\theta_{IM}}$ \cite{dinh2014nice,ardizzone2018analyzing} conditioned on the cancer type label, $y$.  When optimized correctly, the encoded distribution $q(\bm{z^{*}_{M}}|M)$ will match the distribution of the shared latent space $p(f_{\theta_{IM}}(\bm{z^{*}_{I}}|y))$.  Intuitively, we are taking advantage of the invertibility of the NF architecture to let the cancer label type guide the distribution of the image domain to the distribution of the mutation or vice versa.

We can then predict a somatic mutation profile using these steps:
\begin{enumerate}
  \item Input the lesion point cloud into $X_I$ to generate $\bm{z_I}$, which we use to create $\bm{z^{*}_{I}}$
  \item The network of the shared latent space $f_{\theta_{IM}}^{-1}$ given the cancer type label is then used to map $\bm{z^{*}_{I}}$ to $\bm{z^{*}_{M}}$ 
  \item Generate domain specific latent space $\bm{z_{M_0}}$ conditioned on $\bm{z^{*}_{M}}$ (from the above step) via the trained conditional NF prior $p(\bm{z_{M_0}}|\bm{z^{*}_{M}})$ 
  \item Generate the parameters $\bm{r_M}$ and $\bm{p_M}$ of the NB likelihood using $\bm{z_M} = [\bm{z_{M_0}},\bm{z^{*}_{M}}]$ from the previous steps
  \item Predict the mutation profile $M_n \sim NB(\bm{r_M}, \bm{p_M})$
\end{enumerate}
Since the two domains are independent with respect to the latent spaces, we can summarize the generative model above as a joint distribution and factorize:
\begin{equation}
\begin{gathered}
p(M, I, \bm{y}, \bm{z_{I_0}},\bm{z_{M_0}},\bm{z^{*}_{I}},\bm{z^{*}_{M}}) = \\
p_{\theta_{M}}(M| \bm{z_{M_0}},\bm{z^{*}_{M}})p_{\theta_{I}}(I|\bm{z_{I_0}},\bm{z^{*}_{I}}) \\ p_{\eta_{I}}(\bm{z_{I_0}}|\bm{z^{*}_{I}})p_{\eta_{M}}(\bm{z_{M_0}}|\bm{z^{*}_{M}})p_{\theta_{IM}}(\bm{z^{*}_{I}}|\bm{y})p_{\theta_{IM}}(\bm{z^{*}_{M}}|\bm{y})
\end{gathered}
\end{equation}
where $p(\bm{z_{I_0}}|\bm{z^{*}_{I}})$ and $p(\bm{z_{M_0}}|\bm{z^{*}_{M}})$ are the conditional NF priors.  The distributions, $p(\bm{z^{*}_{I}}|\bm{y})$ and $p(\bm{z^{*}_{M}}|\bm{y})$, are of the shared latent space.  The subscripts $\eta$ and $\theta$ indicate the distinct network parameters for each domain.  Similarly, we can factorize the approximated latent posterior:
\begin{equation}
\begin{gathered}
q(\bm{z^{*}_{M}},\bm{z^{*}_{I}},\bm{z_{I_0}},\bm{z_{M_0}}|M,I) = \\ q_{\phi_{M}}(\bm{z_{M_0}}|M)q_{\phi_{I}}(\bm{z_{I_0}}|M) \\
q_{\phi_{M}}(\bm{z^{*}_{M}}|M) q_{\phi_{I}}(\bm{z^{*}_{I}}|I) 
\end{gathered}
\end{equation}
Note that the cancer type label, $y$, is not involved in the latent posterior approximation since it is not involved in the encoder architecture of the model.

\subsection{Variational Objective}

We optimize the neural network parameters by  maximizing the Evidence Lower Bound (ELBO) with encoder and decoder parameters $\theta$ and $\phi$ respectively:

\begin{equation}
  \begin{gathered}
       ELBO(\bm{\Omega}) = \\ \Exc_{p_{f_{\theta_{IM}}}(\bm{z^{*}_{M}}|y)q_{\phi_M}(\bm{z_{M_0}}|M)}[\log_{\theta^{r},\theta^{p}}(M|\bm{z_{M_0}},\bm{z^{*}_{M}})] \\
      \Exc_{p_{g_{\theta_{IM}}}(\bm{z^{*}_{I}}|y)q_{\phi_I}(\bm{z_{I_0}}|I)}[\log_{\theta^I}(I|\bm{z_{I_0}},\bm{z^{*}_{I}})] \\
      + \mathcal{L}_{\bm{z^{*}_{M}}} + \mathcal{L}_{\bm{z^{*}_{I}}}  \\
      -KL[q_{\phi_M}(\bm{z_{M_0}}|M,\bm{z^{*}_{M}})|p_{\eta_{M}}(\bm{z_{M_0}}|\bm{z^{*}_{M}})] \\
      - KL[q_{\phi_I}(\bm{z_{I_0}}|I,\bm{z^{*}_{I}})|p_{\eta_{I}}(\bm{z_{I_0}}|\bm{z^{*}_{I}})]
  \end{gathered} 
  \label{eq:elbo}
\end{equation} 
Thus, the overall objective can be considered a hybrid of the standard VAE objective and maximum likelihood estimation (MLE) with respect to the neural network parameters \bm{$\Omega$}.  The first two terms in Equation \ref{eq:elbo} are the reconstruction errors of the original data.  The middle two terms are losses of matching the shared latent space with latent space from the encoder discussed in Section \ref{optlabel}.  Lastly, the KL divergence of the domain specific latent space is a MLE with respect to the conditional NF prior with an entropy regularizer as shown in Equation \ref{eq:klpr}.  Using Equation \ref{eq:nf} we can rewrite the KL divergence for the lesion domain as:
\vspace{-.28 cm}
\begin{equation}
  \begin{gathered}
      KL[q_{\phi_I}(\bm{z_{I_0}}|I,\bm{z^{*}_{I}})|p_{\eta_{I}}(\bm{z_{I_0}}|\bm{z^{*}_{I}})] = \\ -\Exc_{q_{\phi_I}(\bm{z_{I_0}}|I)}[p_{\epsilon}(f_{\eta_{I}}(\bm{z_{I_0}}|\bm{z^{*}_{I}})) \\ + \log(\text{det}(\frac{df_{\eta_{I}}}{dz}))]  
      + \mathcal{H}(q_{\phi_I}(\bm{z_{I_0}}|I)).
  \end{gathered} 
  \label{eq:klpr}
\end{equation}

The last term, $\mathcal{H}$, is the entropy of the approximate posterior distribution from the encoder.  The first and second terms are the log-likelihood of $z_{M_0}$ under the prior distribution modeled by the conditional NF.  Note that we can simply sample from $q_{\phi_I}(\bm{z_{I_0}}|I)$ to calculate the entropy.

The full learning paradigm is visualized in Figure \ref{fig:fullModel}.  To learn the parameters of the encoder, decoders, and the latent space we optimize the negative of Equation \ref{eq:elbo} via bidirectional training.  Before we update any parameters, we calculate the loss of the invertible networks in the forward and reverse directions given a sample from both domains.  This technique encourages both domains to influence the parameters of the shared latent space and the learnable priors.

\subsection{Optimization of Shared Latent Space}\label{optlabel}

To encourage sharing of information, we want the shared latent space, $p(\bm{z^{*}_{I}}|y)$, to match the generated latent space $q(\bm{z^{*}_{I}}|I)$ after the flow transformation $f_{\theta_{IM}}^{-1}$.  Instead of using the KL divergence to optimize the loss, we use the Maximum Mean Discrepancy (MMD) for $\mathcal{L}_{\bm{z^{*}_{M}}} \text{and } \mathcal{L}_{\bm{z^{*}_{I}}}$ .  The MMD divergence confers two advantages.  First, we can explicitly fit the distribution of $\bm{z^{*}_{I}}$ without any assumptions about a prior.  Secondly, if convergence is reached, we can sample from the shared latent space after the respective flow transformation and disregard the need to generate random samples from a fixed prior during test time.  The only requirement is that the size of $|\bm{z^{*}_{M}}|$ matches the size the $|\bm{z^{*}_{I}}|$.  This optimization procedure is similar to ones used in Adversarial Autoencoders \cite{makhzani2015adversarial}, where instead the Jensen Shannon divergence is used as it scales to higher dimensions.  Since we can control the size of the shared latent space, we are not limited by the scaling issues of MMD.  So we can conceptualize this as mapping the shared latent space to domain specific embeddings of the somatic mutation profile.  

\section{Experiments}

We use 70\% of the dataset to train, 15\% to validate, and 15\% to test our model for accurate reconstruction and prediction of a somatic mutation profile.  Further implementation details are discussed in the Supplemental Material (Appendix \ref{implement}).  

We compare two versions of LLOST, one with a NB likelihood, LLOST\textsubscript{NB}, and another with the Bernoulli likelihood, LLOST\textsubscript{B}, along with the three baseline CVAE models.  For the baseline models, $\text{CVAE}_{r},$ uses the ResNet \cite{he2016deep} architecture to extract imaging features from a single lesion slice with a NB likelihood decoder. The $\text{CVAE}_{p}$ and $\text{CVAE}_{pb}$ models both use the point cloud architecture from our model, but $\text{CVAE}_{p}$ has an NB likelihood decoder and $\text{CVAE}_{pb}$ has a Bernoulli likelihood decoder.   We then follow the CVAE methodology by concatenating both the somatic mutation profile matrix and the cancer type matrix with the output of the encoders and decoders.  The baseline models use a NF prior for the latent space.   Testing is done by generating the latent embeddings for the lesion and then concatenating it with random samples from the latent space.  

Our experiments on the shared latent space are done via an ablation study, where we remove the conditioning by the cancer type label.  We also study the effect of increasing the width of the shared latent space.  

\subsection{Comparison Metrics}

We use log perplexity, $-\frac{1}{N}\Sigma^{N}_n\frac{1}{L_n}\log p(M_n|z_{m_0},z^{*}_{m})$, to compare the goodness-of-fit of our models and the baseline models evaluated over 35 epochs.  The perplexity metric is a function of the reconstruction error, $\log{ p(M_n|z_{m_0},z^{*}_{m})}$, total number of mutations in a sample, $L_n$, and the number of the samples, $N$.  

We use the root mean squared error (RMSE) and TML to measure the prediction error of the models with NB likelihoods.  The RMSE shows the average error for the count of mutations for each distinct gene.  TML is a relatively recent biomarker used for determining sensitivity to targeted treatments \cite{chalmers2017analysis} and is simply the sum of all mutations within a sample.  We calculate the prediction error of TML using a simple point estimate, $\text{TML}^{*}- \text{TML}$, where $*$ indicates our predicted value.

We use the F1-score and Positive Predictive Value (PPV) metrics to assess the Bernoulli likelihood model.  Specifically, we use the F1-score to compare the experiments on the shared latent space.  The PPV metric is used to determine the predictive accuracy of LLOST\textsubscript{B} for specific cancer types. To calculate this metric we use 15\% of the samples of a specific cancer type for testing, while the remaining samples are used for training.

\begin{figure}[h!]%
\centering
\includegraphics[width=1\linewidth]{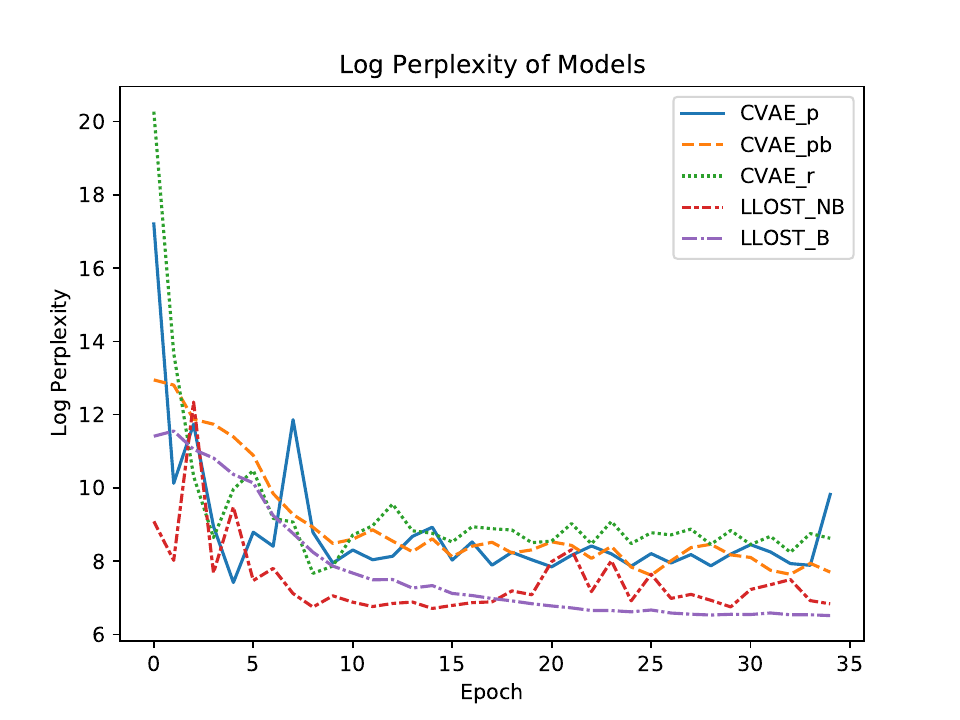}%
\caption{Log Perplexity  (lower is better) of CVAE\textsubscript{p}, CVAE\textsubscript{pb}, CVAE\textsubscript{r}, LLOST\textsubscript{r}, 
and LLOST\textsubscript{B}as a function of epochs} 
\label{fig:perpmodels}
\end{figure}

\begin{figure*}[t]
\centering
\includegraphics[height=.5\linewidth]{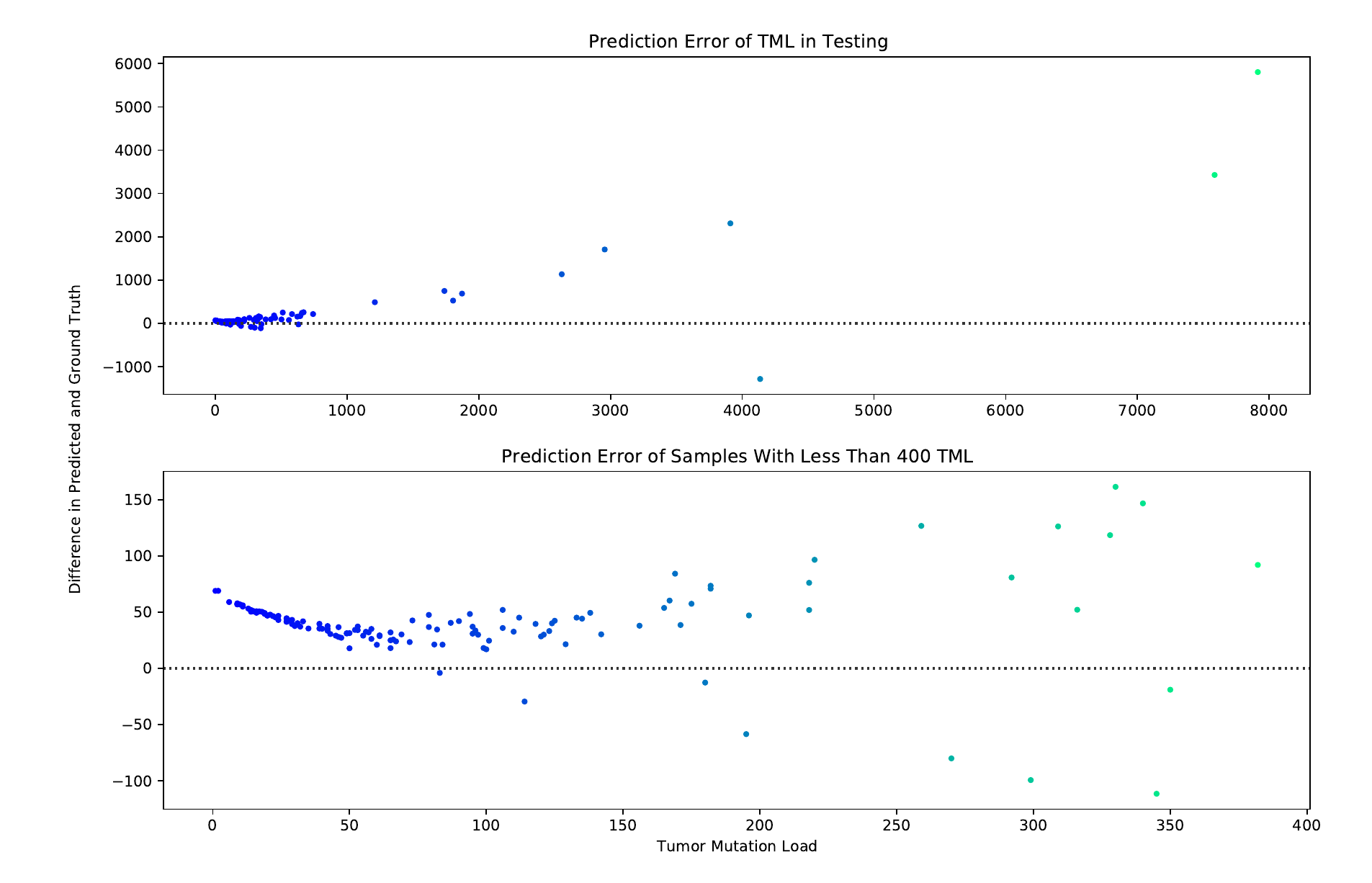}%
\caption{Point prediction error of TML.  The top plot shows the point estimate error in predicting the TML using LLOST\textsubscript{NB} in the test samples.  The bottom plot is a zoomed in of the top plot, where samples with less than 400 TML is reported. X-axis is the expected TML. Y-axis is the difference in TML of predicted and expected.}
\label{fig:tmlrecon}%
\end{figure*}

\subsection{Results}
Figure \ref{fig:perpmodels} displays the overall log-perplexity scores over 35 epochs, whereafter the models are asymptotic and in the case of CVAE\textsubscript{p}, the model begins to collapse.  We observe that LLOST has the best performance in predicting the original somatic mutation profile.  Specifically, LLOST\textsubscript{B} is the best in recreating the original distribution of the somatic mutation profile. This also shows that learning the occurrence of a specific mutation is much easier compared to learning the count of a specific mutation, since CVAE\textsubscript{pb} also out-performs the other CVAE baselines, which have a NB likelihood.  When comparing the CVAE\textsubscript{p} and CVAE\textsubscript{r}, we see that the  point cloud features provide additional information that aids the model to predict the counts of each mutation.  The key insight is that LLOST captures domain specific information and non-trivial shared patterns between the image and genetic domains that reflect cancer type.  

In Table \ref{table:rsme} we show the RMSE of the models with NB likelihoods.  Once again, the CVAE models do not have enough representation power in their latent space to accurately describe the underlying distributions of the two domains.  As the CVAE is simply a concatenation of the encoder with the somatic mutation matrix, the approximate posterior distribution cannot scale to incorporate such a high dimensional dataset.  Although LLOST\textsubscript{NB} still has high RMSE, there is over a 50\% increase in performance of predicting the TML.  This supports our hypothesis that learning distinct and shared latent spaces for very different domains is beneficial for high dimensional prediction tasks.  
\vspace*{-3mm}
\begin{table}[h]
\centering
\scalebox{0.8}{
\begin{tabular}{@{}ll@{}}
\toprule
Model & RSME                 \\ 
\midrule
CVAE\textsubscript{r} & 855.45 \\
CVAE\textsubscript{p} & 731.37 \\
LLOST\textsubscript{NB}  & 315.15                      \\ 
\bottomrule
\end{tabular}}
\caption{RSME of Models with the NB likelihood}
\label{table:rsme}
\end{table}
\vspace*{-3mm}

We examine why the LLOST\textsubscript{NB} has a high RMSE in Figure \ref{fig:tmlrecon}.  We observe the NB likelihood favors over-counting mutations.  The mean prediction error of TML is 54 in samples with less than 400 TML.  When qualitatively assessing the predicted somatic mutation profiles, we observe that genes with the most frequent mutations are predicted to occur almost at least once such as TP53, KRAS, and BRCA1. This pattern continues and is abundantly apparent when TML is larger than 1000, such that commonly occurring mutations occur at least once.  

Overall the LLOST\textsubscript{B} has a lower false positive rate in comparison to LLOST\textsubscript{NB}, which demonstrates the Bernoulli likelihood is not as influenced by the frequency of mutations. This is attributed to parameterization of the Bernoulli likelihood via the NB parameters (Equation \ref{eq:nb_bin}). Since frequently occurring mutations have a significantly lower variance $(p-value < 0.01)$, they will only have a high probability when the  mutation is actually present in a somatic mutation profile.

We observe in Table \ref{tab:cancer_result} the PPV of individual cancer types using  LLOST\textsubscript{B}.  As expected the cancers with a higher number of samples are better able to predict a somatic mutation profile from a corresponding medical images of a lesion.  This table also demonstrates that LLOST\textsubscript{B} does not overfit or underfit, since the PPV varies across all cancer types.  PPV scores for some cancers with lower samples (LUAD, LUSC, COAD, UCEC, and CESC) are higher than expected.  This highlights that the shared latent space in LLOST\textsubscript{B} can learn non-linear many-to-many mappings between the lesion image and somatic mutation domain, as cancers share similar mutations.   

\begin{table*}
\centering
\scalebox{0.9}{
\begin{tabular}{|l|l|l|}
\hline
Cancer Type                                  & Total Samples (15\%) & PPV \\ \hline
Bladder Cancer (BLCA)                        & 119 (18)    & $0.805 \pm 0.117$       \\ \hline
Breast Cancer (BRCA)                         & 139 (21)    & $0.834 \pm 0.319$       \\ \hline
Cervical Squamous Cell Carcinoma (CESC)      & 54 (8)     & $0.726 \pm 0.172$       \\ \hline
Colorectal Cancer (COAD)                     & 21 (3)      & $0.710 \pm 0.067$       \\ \hline
Esophageal Carcinoma (ESCA)                  & 16 (2)      & $0.513 \pm 0.196$       \\ \hline
Glioblastoma Multiforme (GBM)                & 91 (14)      & $0.645 \pm 0.043$       \\ \hline
Head and Neck Cancer (HNSC)                  & 159 (24)     & $0.840 \pm 0.124$       \\ \hline
Kidney Chromophobe (KICH)                    & 15 (2)      & $0.264 \pm 0.015$       \\ \hline
Kidney Renal Clear Cell Carcinoma (KIRC)     & 186 (28)     & $0.829 \pm 0.175$       \\ \hline
Kidney Renal Papillary Cell Carcinoma (KIRP) & 34 (5)      & $0.565 \pm 0.293$       \\ \hline
Brain Lower Grade Glioma (LGG)               & 110 (17)     & $0.732 \pm 0.023$       \\ \hline
Liver Hepatocellular Carcinoma (LIHC)        & 95 (14)      & $0.776 \pm 0.149$       \\ \hline
Lung Adenocarcinoma (LUAD)                   & 41 (6)      & $0.719 \pm 0.271$       \\ \hline
Lung Squamous Cell Carcinoma (LUSC)          & 36 (5)      & $0.725 \pm 0.043$       \\ \hline
Ovarian Serous Cystadenocarcinoma (OV)       & 103 (15)     & $0.749 \pm 0.386$       \\ \hline
Prostate Adenocarcinoma (PRAD)               & 14 (2)      & $0.394 \pm 0.039$       \\ \hline
Stomach Adenocarcinoma (STAD)                & 46 (7)     & $0.672 \pm 0.086$       \\ \hline
Uterine Corpus Endometrial Carcinoma (UCEC)  & 63 (10)     & $0.713 \pm 0.261$       \\ \hline
\end{tabular}}
\caption{Positive predictive value of predicting if a distinct Somatic Mutation occurs in the specific cancer types. The number of test samples for each cancer type is indicated in Column 2 within the parentheses.  The PPV is the mean Positive Predicted Value with standard deviation indicated by $\pm$ using bootstrapping with a 1000 runs.}
\label{tab:cancer_result}
\end{table*}

\begin{figure}[h]%
\centering
\includegraphics[height=.75\linewidth]{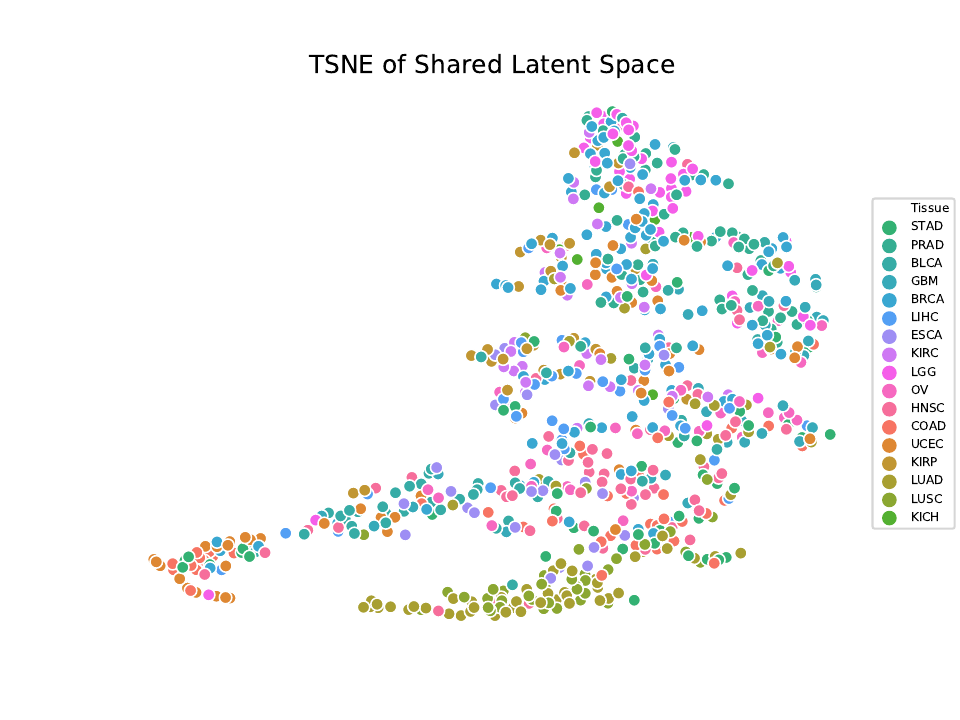}%
\caption{A TSNE of the shared latent space in the forward direction after a test batch of lesions point clouds. Best viewed digitally.}
\label{fig:shared}%
\end{figure}
Figure \ref{fig:shared} displays the shared latent space in the direction towards the mutation domain.  We see that some of the cancers overlap such as LUAD with LUSC and UCEC with COAD.   This is also evident in cancer biology as LUAD and LUSC share similar somatic mutations, where LUSC tends to have a higher TML \cite{bailey2018comprehensive}.  Similarly, cancer biology indicates UCEC and COAD also share somatic mutations \cite{bailey2018comprehensive}.  From the clustering of the shared latent space, we can see that LLOST\textsubscript{B} maps features from the lesion domain by orienting them to match the features of the somatic mutation domain.  In addition we show in the Supplemental Figure \ref{fig:boxplot} the cancer type label is influential to regularize the loss function to optimizing the sharing o the shared latent space. 

\section{Discussion}
Overall, our results indicate that a distinct shared latent space greatly improves the learning of two very different datasets, especially when one dataset is sparse and high-dimensional.  Here we further investigate some of our design choices.

LLOST\textsubscript{NB} generally overestimates the count of somatic mutations.  We hypothesize this is because of an imbalanced dataset where cancer types with larger samples propagate the phenom of cross-excitation \cite{zhou2013negative}. Specifically, if two mutations co-occur together frequently, then the presence of one will excite the other if one is not present.  One method of decreasing this phenom, while not trivial, is to increase the dataset size, so that the conditional probabilities of co-occurrance decrease.  Another avenue, is to change the likelihood to a Zero-Inflate Negative Binomial distribution, which has a superior performance with datasets containing many zeros (absence of mutations).

A major benefit of our model is the use of a distinct shared latent space as it allows us to move from the imaging domain to the mutation domain.  In our original hypothesis, we stated we can use domain specific features along with partial correspondence from the shared latent space to improve prediction accuracy.  We observe this in the high PPV score of COAD even with limited samples.  This is due to the shared somatic mutations between COAD, UCEC, and BLCA, as reported by Bailey et al. \cite{bailey2018comprehensive}, thereby highlighting the influence of domain specific features.

The influence of the shared space is specifically observed in LUAD and LUSC, which both have less than 40 samples.  Although, LUAD and LUSC share very similar somatic mutation profiles, the geometric properties from the lesion domain in conjunction with the cancer type allow the model to discriminate between the two cancer types.  We also observe this again in the uterine based cancers UCEC and COAD.  The geometric properties between these two cancers allows the model to discriminate between their corresponding somatic mutation profiles.

It is also possible to use LLOST\textsubscript{B} to aid clinicians, since somatic mutations themselves are a strong genetic marker for patient prognosis, subtyping, and treatment planning.  For example, BRAF is a somatic mutation targeted in metastatic colon cancer, \cite{ducreux2019molecular}, which our model predicted in COAD with a F1 score of 0.760\footnote{Note this is not the overall F1 score for all mutations in COAD.}.  At the initial diagnosis, the cancer may not seem metastatic, but with the aid of, LLOST\textsubscript{B}, the indication of the BRAF mutation could suggest the clinicians to focus on specific treatment plans that target aggressive types of COAD.  

By predicting a full mutation profile, LLOST\textsubscript{B} can also aid clinicians in determining ineffective treatment plans.  For example, the occurrence of TRIM27 and EGFR together in LUAD is associated with poor response to anticancer therapy in EGFR-mutated lung cancers \cite{balak2006novel}.  Among the LUAD samples, our model predicts TRIM27 and EGFR with an F1 score of .791 in the full mutation profile, thereby these patients could be recommended to receive a different treatment.

There are a number of avenues we can take on improving this model.  A short term goal is to determine a more effective optimization strategy for the shared latent space, such as including a loss to model reconstruction of the cancer type label in the shared latent space.  This could however, overfit, to cancer with higher samples.  A clear extension to this model is incorporation of established radiological and imaging texture features or other genetic domains such as RNA-SEQ.  Neither of these are trivial extensions as it requires a significant change to neural network architecture as we must account for the unordered nature of point clouds and manipulation of the shared latent space.  The flexibility of LLOST, however, allows for an increased dataset without increased computational power.  This will increase confidence in predictive performance and potentially allow clinicians to quickly determine an effective treatment plan for cancer patients.
\section{Conclusion}
We presented, LLOST, a deep latent variable model for the prediction of a somatic mutation profiles of patients based on their corresponding image using dual variational autoencoders joined together with a separate latent space.  We have shown that it is possible to predict somatic mutation profiles without reducing the dimensionalality of the dataset.  We also showed that shared correlations between the imaging and mutation domain help predict mutation profiles when there are a limited amount of samples in the training set.  The LLOST\textsubscript{B} in specific has several attractive features: point cloud representation of a lesion, sharing of a latent space across two significantly different domains, a flexible training objective, and prediction of distinct somatic mutations. 

\bibliographystyle{unsrt}
\bibliography{mehta5_BIB}

\begin{appendices}

\section{Supplemental Information}
\subsection{Ablation Study}

\begin{figure}[h!]
\centering
\includegraphics[width=1\linewidth]{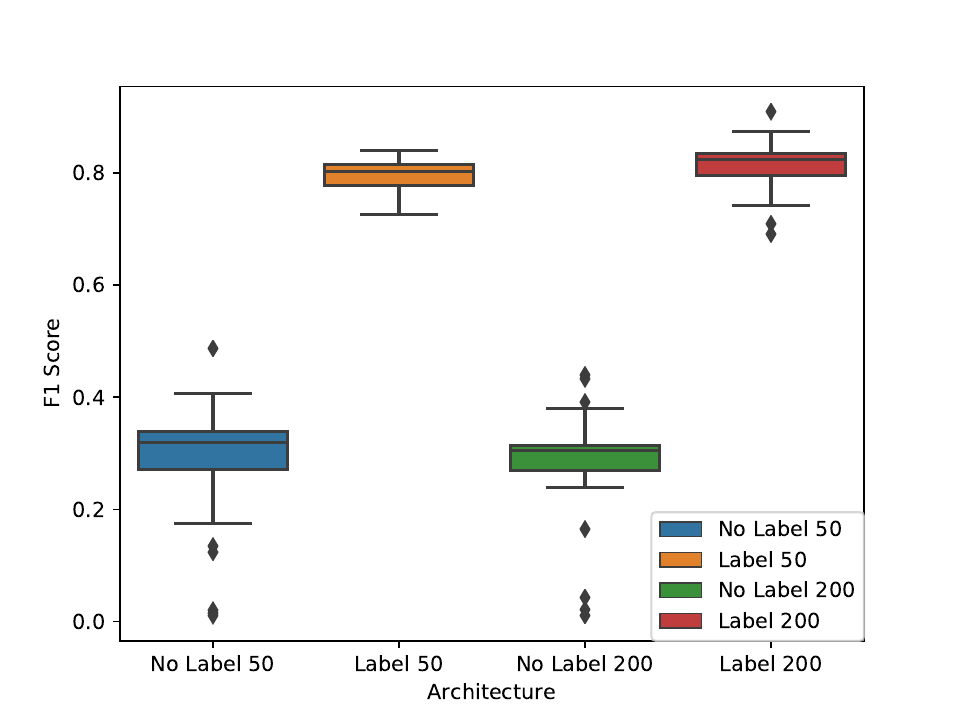}%
\caption{A comparison of the F1 score during testing with two sizes of $z^{*}_{L}$ or $z^{*}_{M}$ and no label in the shared latent space.}
\label{fig:boxplot}
\end{figure}

In Figure \ref{fig:boxplot} we observe that there is no significant difference between the sizes of the shared latent space for predicting the somatic mutation profiles using LLOST\textsubscript{B}, with 50 having .79 and 200 having 0.81 F1-scores.  Furthermore, we can see performance degrades significantly with the ablation study of the cancer type label, $y$ in LLOST\textsubscript{B}.  The label encourages the distribution from the lesion domain to match the structure of the distribution of the mutation domain, akin to supervised machine learning models.   Without the label there are two problems.  First, the trainable prior $p(z_{M_{0}}|z^{*}_{M}$ is using lesion specific features, which leads to a high KL divergence in the both domains.  Secondly, the decoder is essentially using lesion specific features which to it is noise, and therefore the reconstructed distribution is also noisy.

\subsection{Implementation Details}\label{implement}

We adopt the encoder-decoder structure proposed in \cite{achlioptas2017learning} for the point cloud VAE with the size of the latent space $\bm{z_{I}}$ as 512. The encoder-decoder structure of the VAE for somatic mutations is a set of symmetric multilayer perceptron (MLP) with dimensions [1000, 500, 300] when using a NB-likelihood and [800, 500] for the Bernoulli likelihood, where the last dimension indicates size of the embedding space, $\bm{z_{M}}$.   

\begin{figure}[b!]%
\centering
\includegraphics[width=.65\linewidth]{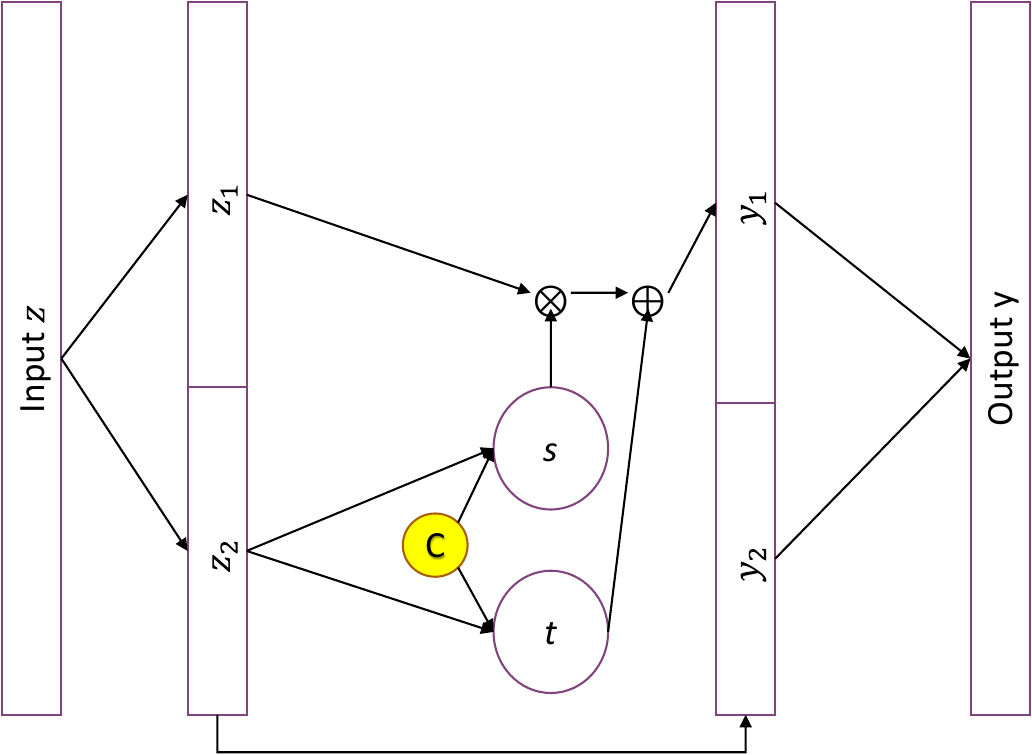}%
\caption{An example of a conditional coupling block where $s$ and $t$ are neural networks.  Here $c$ is the condition that is simply concatenated to each neural network.}
\label{fig:ccblock}%
\end{figure}

Our model architecture for the latent spaces follows the NF architecture of realNVP \cite{dinh2016density}, where one flow with a single affine coupling block for brevity is shown in Figure \ref{fig:ccblock}. The input to the coupling layer is the partition of the input $z = [z_1, z_2]$.  The output of one coupling layer is then $y_2 = z_2$ and $y_1 = z_1 \otimes exp(s(z_2,c)) + t(z_2,c))$ where functions $s$ and $t$ are neural networks, and $c$ is the specific condition.  Conditioning $c$ onto $s$ and $t$ does not affect invertiblility, since the $s$ and $t$ are never inverted, therefore the functions can utilize any neural network architecture. We use 12 flow steps each with two affine coupling blocks for domain specific latent space, and we use 24 flow steps with 3 coupling blocks for the shared latent space.

For the kernel in the MMD loss in Section \ref{optlabel}, we achieved the best results with the inverse multiquadtratic: 
\begin{equation}
  \begin{gathered}
\mathcal{L}_{\bm{z^{*}_{M}}} = k(\bm{z^{*}_{M}}, \tilde{\bm{z^{*}_{M}}}) = 1/(1 + \norm{(\bm{z^{*}_{M}} - \tilde{\bm{z^{*}_{M}}})/h}^2)
  \end{gathered}
  \label{eq:mmd}
\end{equation}
where $\tilde{\bm{z^{*}_{M}}}$ is the distribution from the forward process of $f_{IM}$.  This is also applied in the reverse direction to calculate $\mathcal{L}_{\bm{z^{*}_{I}}}$.

\subsection{Point Cloud Generation}

The code to generate the point cloud examples shown in Figure \ref{fig:interpolated} and in general for any segmented lesion is available at \url{http://github.com/rmehta1987/LLOST}.

\end{appendices}
\end{document}